\documentclass[10pt, conference, compsocconf]{IEEEtran}
\usepackage{graphicx}
\usepackage{cite}
\usepackage{picinpar}
\usepackage{amsmath}
\usepackage{url}
\usepackage{flushend}
\usepackage[latin1]{inputenc}
\usepackage{colortbl}
\usepackage{soul}
\usepackage{multirow}
\usepackage{pifont}
\usepackage{color}
\usepackage{alltt}
\usepackage[hidelinks]{hyperref}
\usepackage{enumerate}
\usepackage{siunitx}
\usepackage{breakurl}
\usepackage{epstopdf}
\usepackage{pbox}
\usepackage{amsfonts}
\usepackage[table]{xcolor}

\usepackage{diagbox}
\usepackage{booktabs,array,xcolor,colortbl,multicol,times,epsfig,graphicx,subfigure}
\usepackage{amsmath,amsthm,amsfonts,amssymb,bm}
\usepackage[vlined,boxed,commentsnumbered,ruled,linesnumbered]{algorithm2e}
\usepackage{caption}
\usepackage{colortbl,accents}
\usepackage{subfigure}
\usepackage{caption}
\renewcommand{\figurename}{Fig.}
\def\widebar{\accentset{{\cc@style\underline{\mskip10mu}}}}

\ifCLASSINFOpdf
\else
\fi
\begin{document}

%
\title{Cognitive Visual Inspection Service for LCD Manufacturing Industry}


\author{\IEEEauthorblockN{Yuanyuan Ding\textsuperscript{1}, Junchi Yan\textsuperscript{2}, Guoqiang Hu\textsuperscript{1}, Jingchang Huang\textsuperscript{1}, Jun Zhu\textsuperscript{1}}
\IEEEauthorblockA{\textsuperscript{1}IBM China Research Lab, Shanghai, China \\
\textsuperscript{2}Shanghai Jiao Tong University, Shanghai, China\\
\{dyuany,hugq,hjingc,zhujun\}@cn.ibm.com, yanjunchi@cs.sjtu.edu.cn
}
}


%


\maketitle

\begin{abstract}
With the rapid growth of display devices, quality inspection via machine vision technology has become increasingly important for flat-panel displays (FPD) industry. This paper discloses a novel visual inspection system for liquid crystal display (LCD), which is currently a dominant type in the FPD industry. The system is based on two cornerstones: robust/high-performance defect recognition model and cognitive visual inspection service architecture. A hybrid application of conventional computer vision technique and the latest deep convolutional neural network (DCNN) leads to an integrated defect detection, classification and impact evaluation model that can be economically trained with only image-level class annotations to achieve a high inspection accuracy. In addition, the properly trained model is robust to the variation of the image qulity, significantly alleviating the dependency between the model prediction performance and the image aquisition environment. This in turn justifies the decoupling of the defect recognition functions from the front-end device to the back-end serivce, motivating the design and realization of the cognitive visual inspection service architecture. Empirical case study is performed on a large-scale real-world LCD dataset from a manufacturing line with different layers and products, which shows the promising utility of our system, which has been deployed in a real-world LCD manufacturing line from a major player in the world.

\end{abstract}

\begin{IEEEkeywords}
LCD; visual inspection; defect classification; defect segmentation

\end{IEEEkeywords}

%
\IEEEpeerreviewmaketitle

\section{Introduction}
The liquid crystal display (LCD) industry is developing quickly recently with the booming market of laptops, monitors, televisions, smart phones, digital cameras and other portable electronic devices. Typical LCD manufacturing can be divided into three stages: array process, cell process, and module process, whereby each stage involves a few sub-stages. Every sub-stage can bring a variety of defects that finally hurt the yield. Early detection and recognition of defects by visual inspection is very important for the product quality control.

Major tasks involved in LCD visual inspection include defect detection, classification and impact evaluation. While defect detection is the fundamental requirement, defect classification plays an important role in evaluating the defect severity as well as identifying the root causes to improve the production procedure. Impact evaluation concerns what kind of malfunction can be caused by a defect, e.g., shorting or cut of circuits.

Traditional manual inspection is labor intensive, time consuming and mental/physical harmful. Consequently, automatic optical inspection (AOI) \cite{Moganti1996Automatic, Chen2008Automatic, Bai2014} is becoming more and more popular. Existing commercial AOI systems are mostly based on traditional machine vision techniques adopting handcrafted application-specific image feature engineering. The performance of resulting algorithms are sensitive to the quality of input images that is influenced by the setup of camera, lighting, etc. As a result, current AOI systems show a tight coupling between image acquisition hardware and software model. Practical AOI system turns up as an integrated device associated with specific inspection station, making it difficult to scale economically. In addition, traditional machine vision algorithms have relatively limited defect recognition capability. Algorithms are generally tuned to reach a high defect detection rate at the expense of a high false alarm rate simultaneously. Manual validation of the detected defects is still mandatory. Besides, current LCD AOI sysems do not support defect classification generally, not mentioning the impact evaluation.

In this paper, we introduce a novel LCD visual inspection solution that is based on two cornerstones: robust/high-performance defect recognition model and cognitive visual inspection service architecture. A hybrid application of conventional computer vision technique and the latest deep convolutional neural network (DCNN) leads to an integrated defect detection, classification and impact evaluation model that can be economically trained with only image-level class annotations to achieve a high inspection accuracy. In addition, thanks to the high modeling competency of the DCNN, the properly trained model is robust to the variation of the image qulity, significantly alleviating the dependency between the model prediction performance and the image aquisition environment. This in turn justifies the decoupling of the defect recognition functions from the front-end device to the back-end serivce, motivating the design and realization of the cognitive visual inspection service architecture.

The remainder of the paper is organized as follows. Defect recognition model is elaborated in Section II. Experiment and model prediction performance is reported in Section III. The visual inspection service architecture is introduced in Section IV. The work is concluded in Section V.


\section{LCD Defect Recognition}

To model the problem, defect detection and classification are relatively straightforward to map to the object detection and classification problem in computer vision. To support an automatic evaluation of the defect impact, a key technical challenge is to precisely extract the defect regions within an input inspection image. This corresponds to the image segmentation problem, i.e., identifying within an image all the pixels associated with a target object.

Recently, DCNN achieved remarkable success in object classification \cite{alexnet, SzegedyCVPR15, inception, Resnet, densnet}, detection \cite{GirshickCVPR14, RedmonCVPR16, LiuECCV16} and segmentation \cite{long2015fully, Ronneberger2015unet, He2017MaskRCNN}. However, most of the DCNN-based models follow a supervised learning approach. The model training requires a large amount of training images labeled with annotations. For classification model, each training image shall be annotated with a class label. For detection, each target object needs to be annotated with a bounding box in training images. For segmentation, all the pixels associated with target objects shall be marked. Although the labeling of all such annotations is tedious labor-intensive work, the per-image class label is regarded as `light-weight' label for the least labeling efforts. Especially, LCD manufacturers generally have a good collection of history inspection images associated with the class annotations. Therefore, in our solution, we considered model training with image-level class labels only. This poses a special challenge for the defect segmentation task.

In computer vision area, training of DCNN-based segmentation models with only image-level class labels have received a lot of attentions in the paradigm of weakly-supervised learning \cite{Durand2017, Wei2017CVPR, wang2018weakly, Zhou2018, Ahn2019CVPR}. Despite of encouraging advancement, the weakly-supervised segmentation models have not yet achieved comparable performance to fully supervised models. Their direct application to visual inspection is limited.

\begin{figure}[!t]
        \centering
        \includegraphics[width=0.44\textwidth]{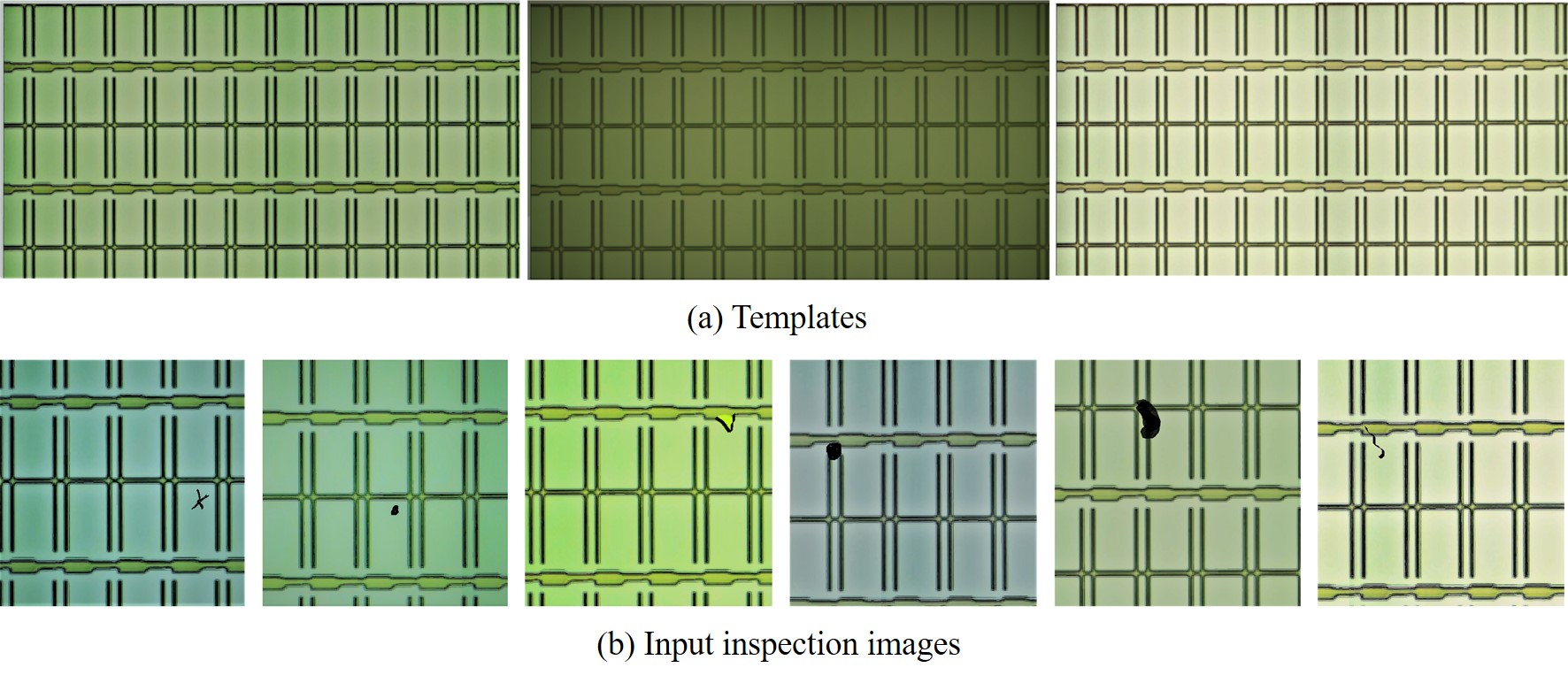}
        \caption{Samples of LCD array inspection image and brightness/hue variation}
        \label{fig:illumination}
\end{figure}

\begin{figure}[tb!]
        \centering
        \includegraphics[width=0.44\textwidth]{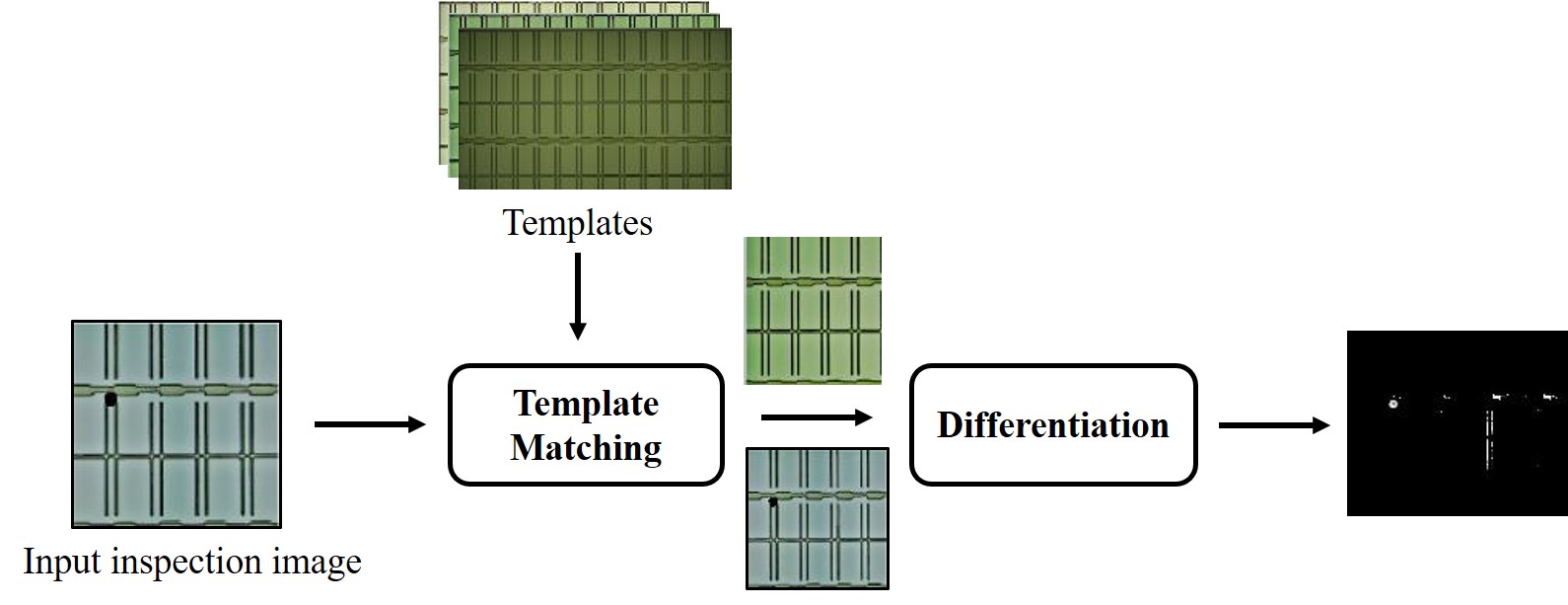}
        \caption{The conventional template matching process.}
        \label{fig:focus}
\end{figure}

Conventional image segmentation techniques, e.g., matching an inspection image with a standard defect-free template, have much less or no reliance on pixel-level annotations. However, its application in the LCD inspection practice is very tricky due to the problem mentioned in Section I. In most cases, because of the variation in the image quality (including brightness, hue, sharpness, etc.), there can be many false alarms by directly applying template matching and differentiation. Fig.~\ref{fig:illumination}(a) shows some pre-designed templates of multiple brightness levels. Fig.~\ref{fig:illumination}(b) illustrates the variety of hue and brightness of inspection images from production lines. Fig.~\ref{fig:focus} shows that the template matching can result in poor segmentation quality. It is clear that there are many false defect areas and it is hard to separate the false ones from the real defect areas.

\begin{figure}[!t]
        \centering
        \includegraphics[width=0.44\textwidth]{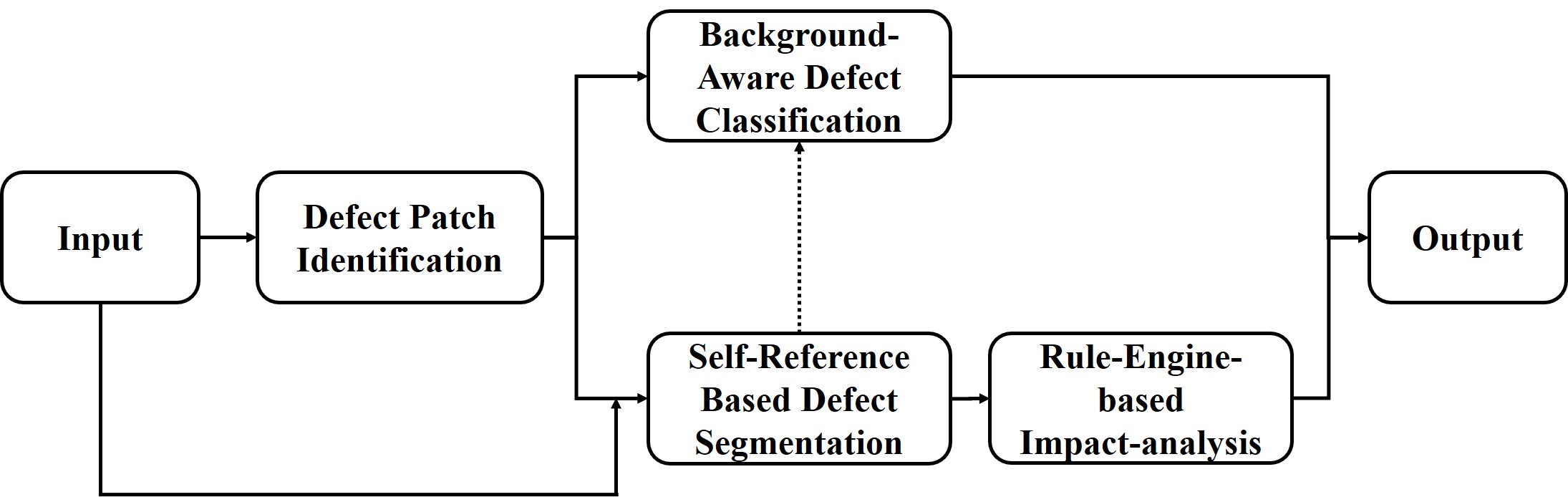}
        \caption{Overview of the complete defect recognition model}
        \label{fig:overall}
\end{figure}

Based on the observation that a LCD inspection image contains a background of periodic patterns (cf. Fig.~\ref{fig:illumination}) and individual defects are generally small, we propose a self-reference based template matching approach for the defect segmentation problem, which is free of training and image annotation. It relies, however, on robust detection and extraction of defect patches (if any) of constant size within inspection images as a first step. This is realized via a DCNN and denoted by the \emph{defect patch identification} module in the overall model architecture of Fig.~\ref{fig:overall}. In addition to the defect area, each defect patch still contains a significant area of defect-free background. Thus, a defect patch can be further matched within the rest part of its containing image for a defect-free correspondence, on the basis of which the defect segmentation can be performed by image subtraction. This is realized by the \emph{self-reference based defect segmentation} module in Fig.~\ref{fig:overall}. Because the image quality (brightness, sharpness, etc.) within the same image is quite homogeneous, the matching and subtraction of the defect patch within the defect-free part of the same image generally results in quite promising defect segmentation. A subsequent \emph{rule-engine-based impact analysis} module evaluates the impact of each defect to the LCD panel according to the defect segmentation results.

For accurate defect classification, we propose a \emph{background-aware defect classification} module (Fig.~\ref{fig:overall}). It is a DCNN-based classifier that inspects not only each defect patch, but also the correspondent defect-free patch obtained for the defect patch as a by-product in the above segmentation task. We verified that this method outperforms the classifier inspecting defect patches only.

In summary, the proposed inspection model combines the advantages of both conventional image processing techniques and deep learning to achieve robust and effective defect classiciation / segmentation, with only image-level class annotations available for the model training. In the following subsections, defect patch identification, self-reference based defect segmentation and background-aware defect classification will be elaborated.

\subsection{Defect Patch Identification}

In our system, the inspection images are first divided into constant-size patches by sliding window. Then, each patch is inspected by a CNN-based binary classifier to determine whether it contains defects or not. To train the classifier, image patches annotated with class labels, i.e., defect or no-defect, are necessary. However, in the raw training data set, a class label is marked for each entire image, which is not sufficient to derive the class labels for individual patches within an image. To tackle with the problem, we jointly applied the weakly supervised learning \cite{Durand2017, Wei2017CVPR, wang2018weakly, Zhou2018, Ahn2019CVPR} and periodic pattern based approach to obtain the patch-level binary class annotations.

\begin{figure}[!t]
        \centering
        \includegraphics[width=0.44\textwidth]{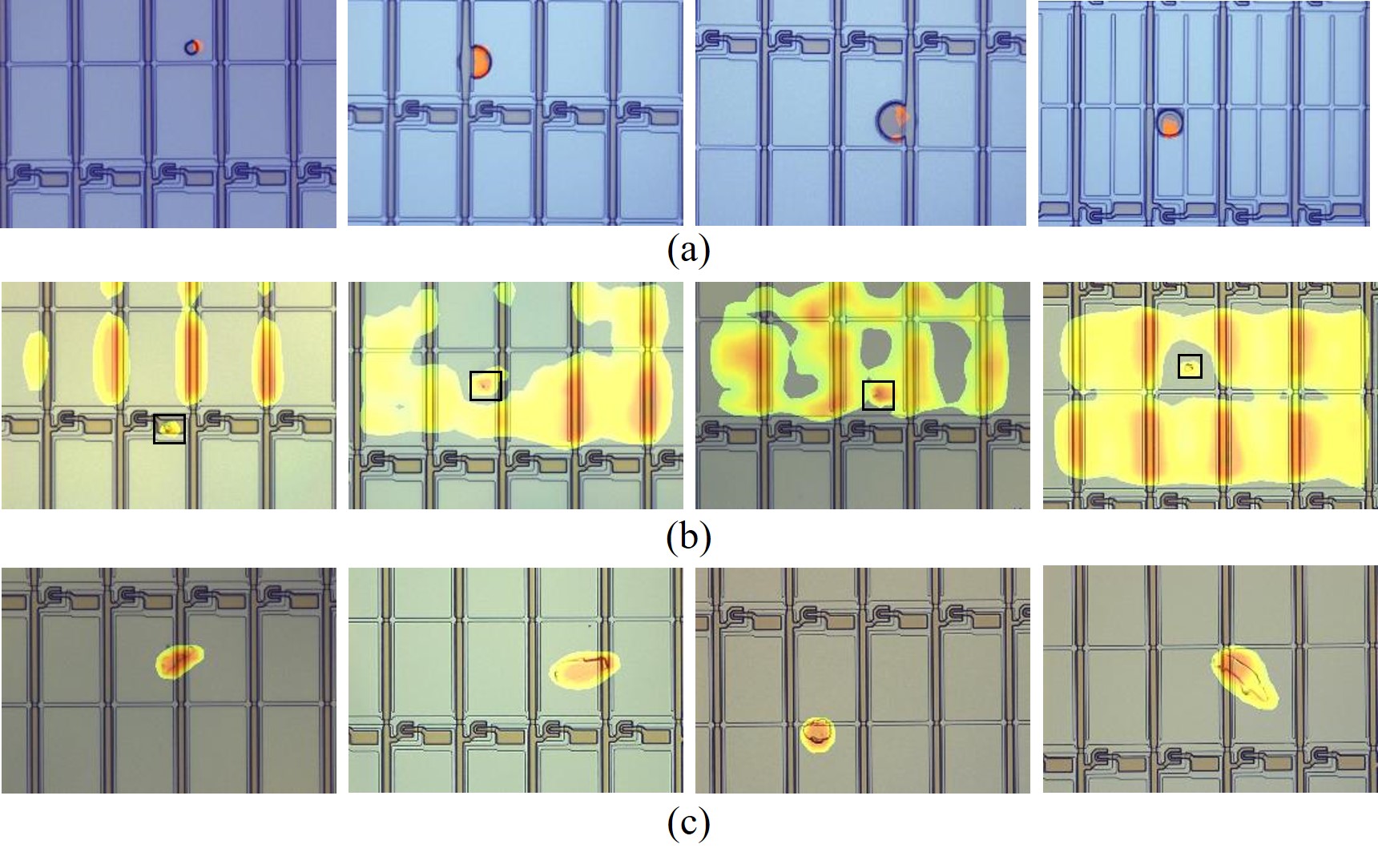}
        \caption{The localization results by the activation heatmap in our experiment. (a) Rough defect regions that are smaller than the real defect areas. (b) Rough defect regions that are larger than the real defect areas. (c) Real defect areas that are well captured.}
        \label{fig:heatmap}
\end{figure}

With weakly supervised learning, we trained firstly a binary classifier to determine whether a raw inspection image (instead of its patches) contains defects or not, with the available inspection images and their class labels.
An activation heatmap \cite{Durand2017, Wei2017CVPR, wang2018weakly, Zhou2018, Ahn2019CVPR} can be obtained from the binary classifier to locate the rough defect regions, which can be further applied to derive whether an arbitrary patch from an image contains defect or not. However, the rough defect regions can be much smaller or larger than the real defect areas. The quality of image patch labeling via only this approach is not sufficient for model training. Fig.~\ref{fig:heatmap} shows some localization results by the activation heatmap method.

\begin{figure}[!t]
        \centering
        \includegraphics[width=0.36\textwidth]{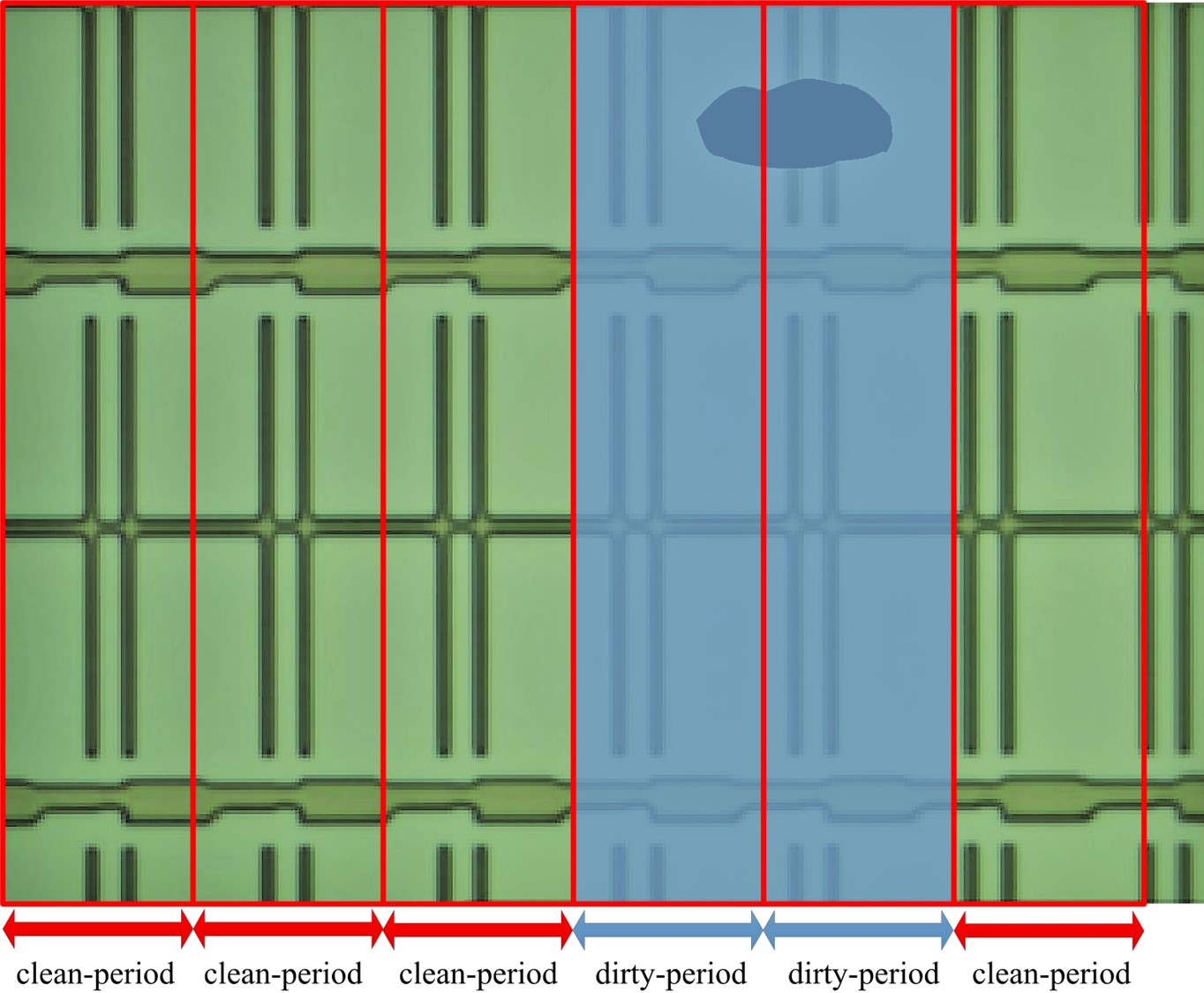}
        \caption{An LCD image with four horizontal periods. The rectangle regions highlighted by blue arrowed lines are corrupted-periods with defect. Other regions are the clean-periods that can be used to construct the referential image.}
        \label{fig:period_est}
\end{figure}

\begin{figure}[tb!]
        \centering
        \includegraphics[width=0.44\textwidth]{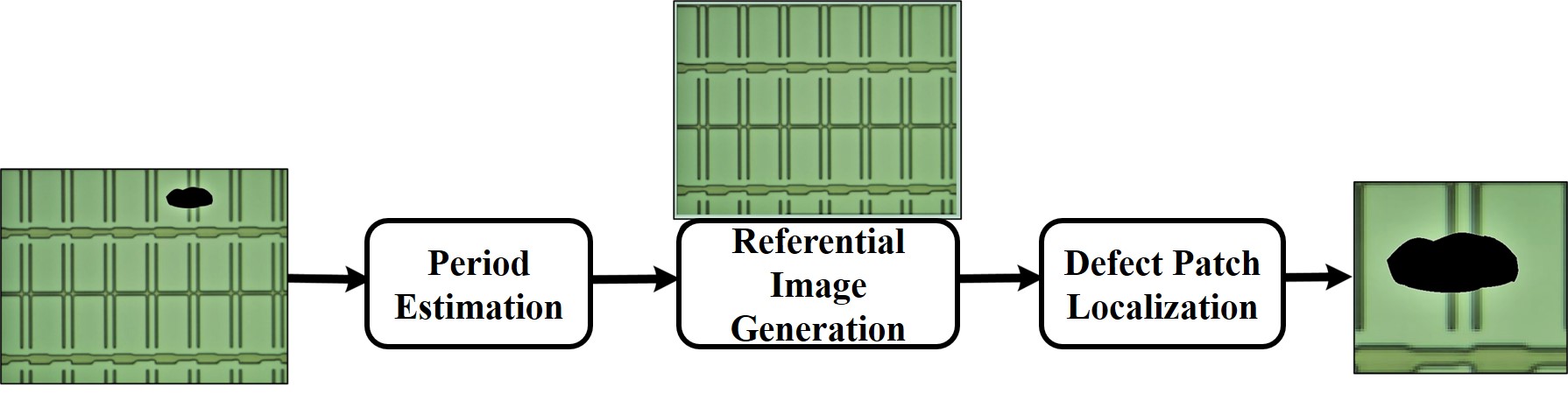}
        \caption{Period pattern based defect patch identification}
        \label{fig:defect_process}
\end{figure}

In addition, we leverage the periodic pattern within the inspection image to identify the number of periods within an image together with a clean period (cf. Fig.~\ref{fig:period_est}) that is not corrupted by any defect. On this basis, a defect-free reference image can be generated by duplicating the clean period. A rough defect area can be further identified by differentiating the inspection image and the generated reference image. The procedure is illustrated in Fig. \ref{fig:defect_process} and explained as follows.

\begin{figure}[tb!]
        \centering
        \includegraphics[width=0.44\textwidth]{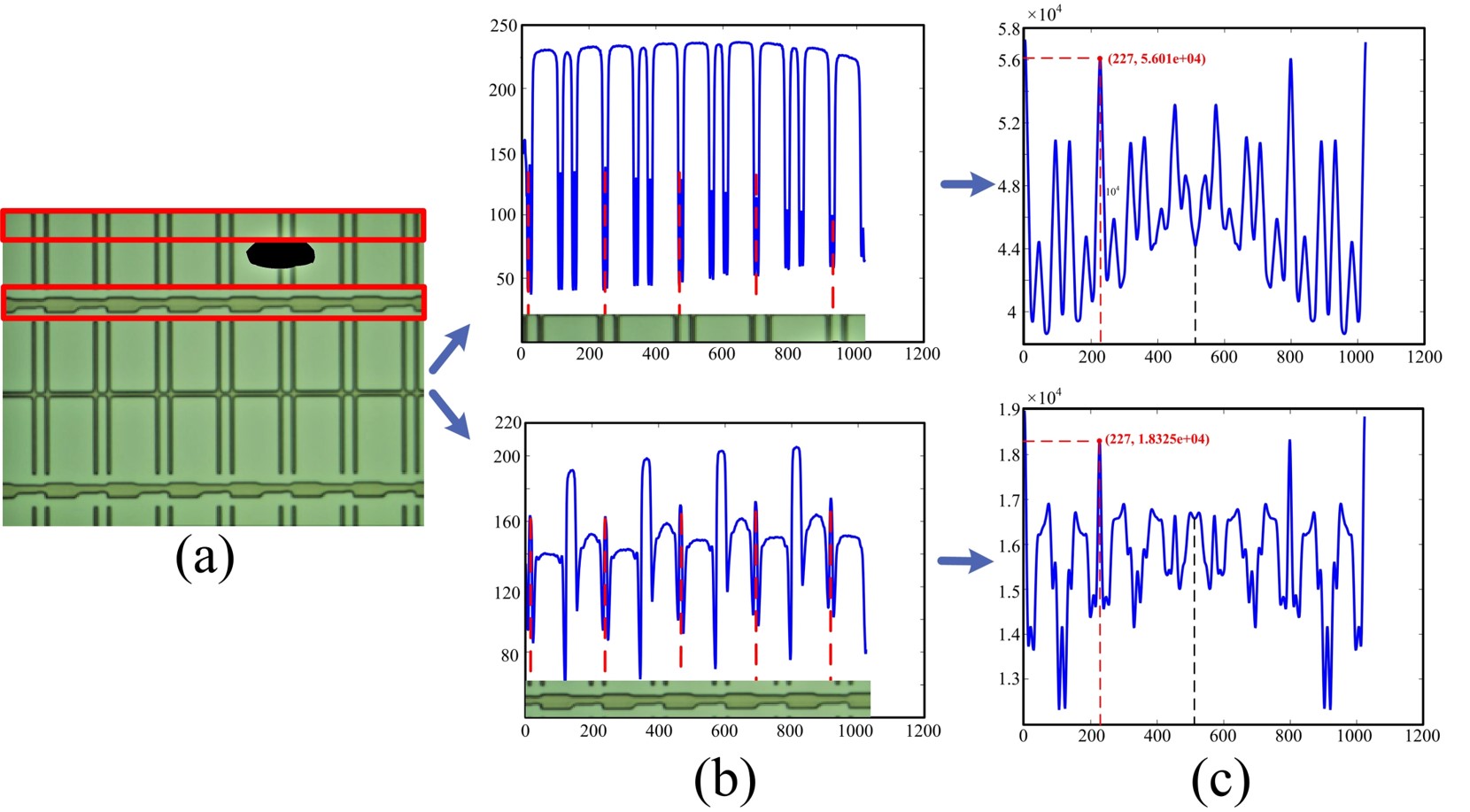}
        \caption{(a) TFT-LCD array image scanned with sub-images (red rectangles). (b) The $\mathbf{P}^k$ curves of the two sub-images. (c) Amplitude curves of the corresponding $\xi_k$.}
        \label{fig:period_window}
\end{figure}

\textbf{Step 1: Period Estimation} Let $I(m,n)\in \mathbb{R}^{M\times N}$ be the (gray) LCD image of size $M\times N$. A set of (horizontal) projections $\mathcal{P}=\{\mathbf{P}^0, \mathbf{P}^1, \ldots, \mathbf{P}^{(M-w)/s}\}$ whose $n$-th element in each $\mathbf{P}^k\in \mathbb{R}^{N}$ for $n\in\{0,1,\ldots,N-1\}$ is defined:
\begin{equation}
\left\{P^k(n)=\sum_{m=ks}^{ks+w}I(m,n)\right\}_{k=0}^{\lfloor\frac{M-w}{s}\rfloor}
\label{eq:proj}
\end{equation}
where $w$ is the window size of the sub-image for projection, and $s$ is the step length. Note we allow for overlapping projection windows hence $s<w$.

Symmetric Average Magnitude Sum Function (SAMSF) \cite{ShahnazTASLP12} is a state-of-the-art method to extract pitch harmonic and the corresponding harmonic number, which has been widely used in speech recognition. We adapt the model to the image domain. The period of the LCD images can be seen as a pitch harmonic of the SAMSF. As a result, the image-domain average magnitude sum function of $P^k(n)$ can be defined as the same as (3) in \cite{ShahnazTASLP12} for $p\in\{0,1,\ldots,N-1\}$:
\begin{equation}
\left\{\xi_k(p)=\sum_{n=0}^{N-1}\|P^{k}(l)+P^{k}(n)\|\right\}_{k=0}^{\lfloor\frac{M-w}{s}\rfloor}
\label{eq:SAMSF}
\end{equation}
where $l=mod(n+p,N)$ and $P^{k}(\cdot)$ is the normalized version by subtracting the mean of $\{P^k\}_{k=1}^{\lfloor\frac{N-w}{s}\rfloor}$ over $k$.

Fig. \ref{fig:period_window} shows an example to estimate the period number and period length over an LCD inspection image with four periods in the horizontal direction together with a dark defect region. The image is then partitioned into a series of sub-images in the vertical direction. Two example defect-free sub-images are depicted by the red rectangles in
Fig. \ref{fig:period_window}(a). Fig. \ref{fig:period_window} (b) shows the corresponding $\mathbf{P}^k$ of the sub-images. The curves in Fig. \ref{fig:period_window} (c) are the amplitudes of $\xi_k$. It is obvious that $\mathbf{P}^k$ well captures the periodicity of the sub-images. Therefore, we can firstly apply SAMSF to obtain the period of $\mathbf{P}^k$ and then estimate the period. The peaks of $\xi_k$ at multiples of the period tend to be well-preserved and remain prominent. On the other hand, in case a sub-image happens to overlap with the defect region, there will be an obivous interruption of the periodicity, the horizontal position of which can also be roughly determined. Based on the processing results of all the sub-images, the clean-/dirty periods (cf. Fig.~\ref{fig:period_window}) of an inspection image can be determined.

The period length $\{T^k\}_{k=0}^{\lfloor\frac{M-w}{s}\rfloor}$ is estimated from each $\{P^k\}_{k=0}^{\lfloor\frac{M-w}{s}\rfloor}$ respectively. Then median filter is applied to find the final estimation $T$ and the corresponding period number $C=\lfloor \frac{N}{T}\rfloor$ in the inspection image. $T^k$ is finally computed by:
\begin{equation}
T^k=\arg\max_{p=\{1,2,\ldots,N-1\}}\xi_k(p)
\label{eq:period_k}
\end{equation}

\textbf{Step 2: Referential Image Generation} Denote the referential template image as $R(m,n)\in \mathbb{R}^{M \times TC}$ which can be reconstructed from the LCD image $I$ by:
\begin{equation}
R(m,n)=I(m,n-\lfloor n/T\rfloor T)
\label{eq:reconstruct}
\end{equation}
In fact $R$ can be a $C$ times copy of an arbitrary clean period identified from the image.

\textbf{Step 3: Defect Patch Localization} We compute the absolute differential value between $R$ and $I$ based on which optional smoothing and morphological filtering can be performed before final binarization. Finally we use a constant-size e.g. $224 \times 224$ box to frame the defect area. These $224 \times 224$ image pathches are the final positive training samples for the binary classifier.

\vspace{\baselineskip}
The periodic pattern based approach has the disadvantage of high false alarm rate, namely, leading to many defect-free patches labeled as defect. Therefore, the weakly supervised approach is further applied to double confirm that a patch should be labeled as defect. This process leads to a set of defect image patches with label accuracy up to $85\%$. Subsequently, a manual screening of the miss-labeled images shall be conducted to clean the data set. Note that the screening operation is just a retrieval of the pre-identified defect patches and removing a small portion of patches that have no defects. The labor cost and time is much less than any of the image annotation procedure for DCNN model training.

While a clean data set of defect patches is collected by the above method, the data set of defect-free patches can be automatically obtained by randomly cutting from those training images free of defects. In this way, an image patch set with high quality labels can be well prepared for the training of the defect patch identidication module. The resulting DCNN classifier generally achieves a high prediction accuracy over $99\%$.

\subsection{Self-Reference Based Defect Segmentation}
For defect impact evaluation, some geometrical checklist needs be fulfilled to see if some key areas or components have been affected by defects, for which pixel level segmentation is necessary. To this end, we propose the self-reference based defect segmentation to get an ideal segmentation result. Instead of matching an inspection image with an offline-collected template image, which is usually very sensitive to image quality and inclined to generate many false alarms, we resort to an adaptive technique that uses an inspection image itself as the reference template.

\begin{figure}[tb!]
        \centering
        \includegraphics[width=0.44\textwidth]{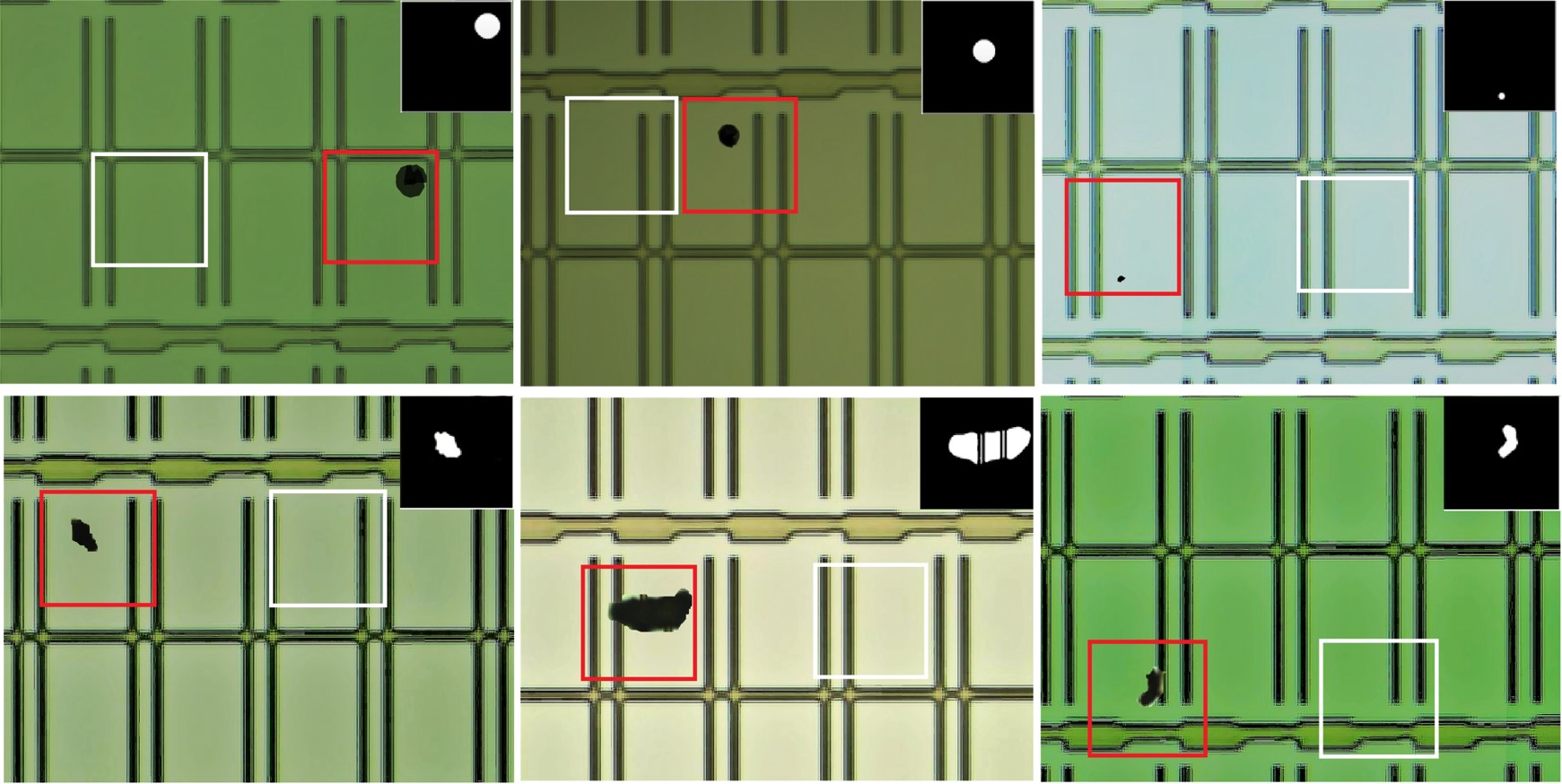}
        \caption{The self-reference based defect segmentation. The red and white bounding boxes respectively indicate the defect and corresponding background patches of $224 \times 224$. The binary defect segmentation images are in the upper right corners of inspection images (size $1024 \times 768$).}
        \label{fig:mulChinnels}
\end{figure}

The approach and segmentation effects are illustrated by six examples in Fig. \ref{fig:mulChinnels}. The red bounding boxes stand for identified defect patches from the preceding stage. Here, the template matching is applied with the defect patch as the target image and the rest defect-free part of the raw image as tempalte. We named it self-reference based tempalte matching method. The white bounding box denotes the matching result which is also regared as the defect-free background of corresponding defect patch. The binary image in Fig. \ref{fig:mulChinnels} is the segmentation result that is crucial for subsequent impact evaluation.

It is clear from Fig.~\ref{fig:mulChinnels} that a good matching of the defect-free background is obtainable from the raw images. The reason is that the raw image contains multiple periods in the horizontal direction and the defect patch also contains a considerable area of the defect-free background with a proper setting of the patch size, leading to a good chance of high quality matching. Besides, the defect and background patches are coming from the same inspection image. So they share the same image acquisition condition. The binary defect segmentation images depict perfect defect areas with negligible noise.

\begin{figure}[tb!]
        \centering
        \includegraphics[width=0.44\textwidth]{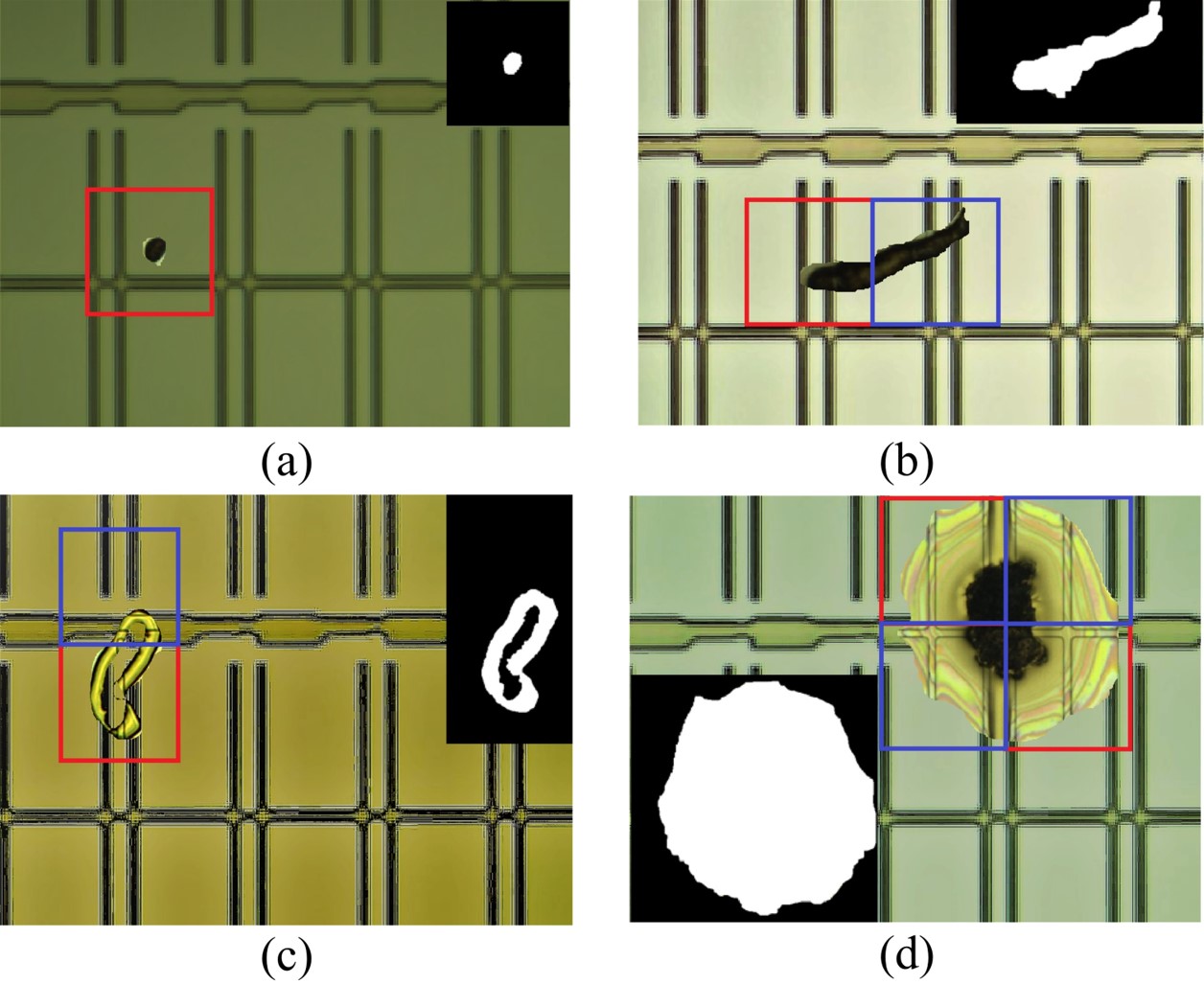}
        \caption{Different region merging results in our case. (a) Defect region patch of $224 \times 224$. (b) Defect region patch of $448 \times 224$. (c) Defect region patch of $224 \times 448$. (d) Defect region patch of $448 \times 448$.}
        \label{fig:regionPolicy}
\end{figure}

Additional region merging policy was adopted to handle the situation where relatively large defects are split into multiple defect patches. Fig.~\ref{fig:regionPolicy} shows some cases in our work. After the merging, the defects of large sizes are successfully recognized as complete ones. Such defects, however, amount to just 5\% of the total defects.

\subsection{Background-Aware Defect Classification}\label{subsec:bdc}

Equipped with the image level defect class annotations together with the defect patch labeling in Section II.A, a training data set of multi-class defect patches can be prepared. On this basis, a defect classifier is trained for defect patches.

\begin{figure}[tb!]
        \centering
        \includegraphics[width=0.44\textwidth]{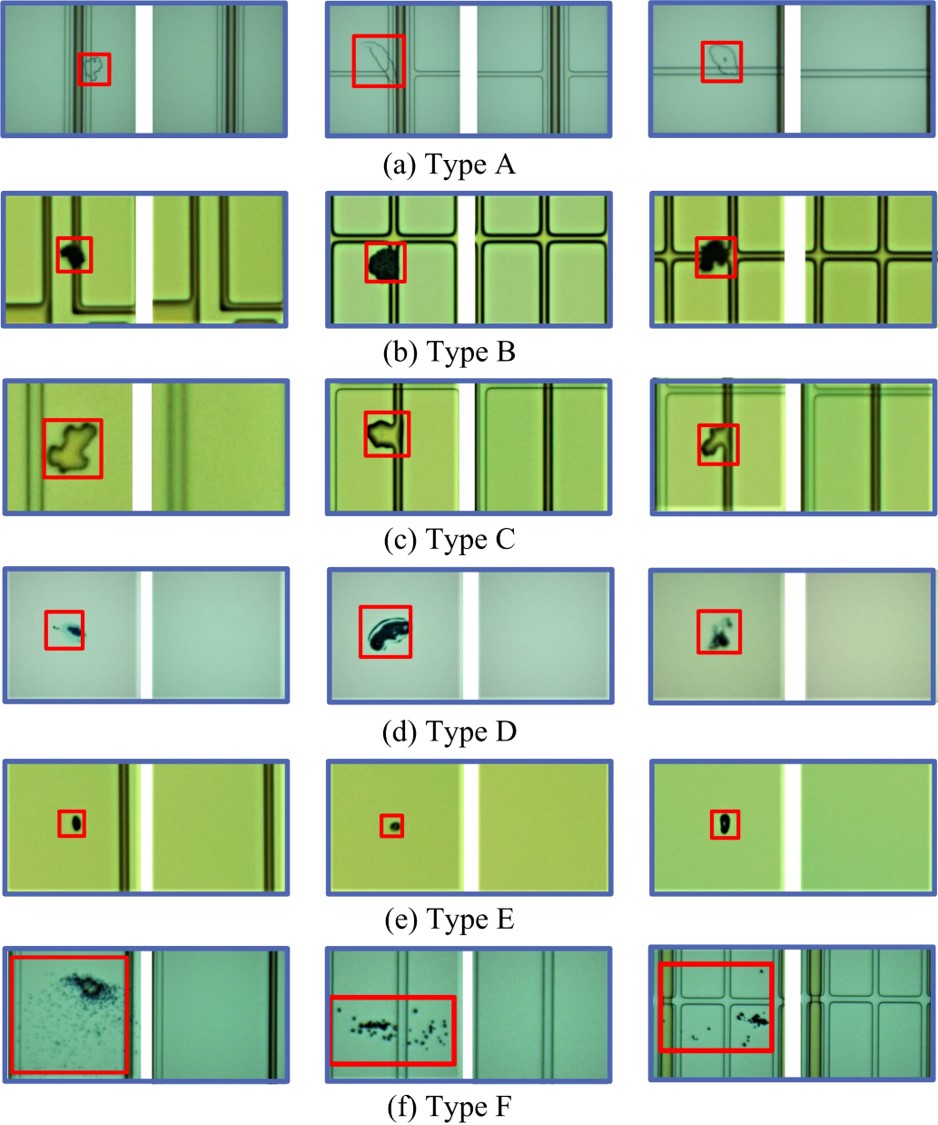}
        \caption{Samples of different types of the defect patches and the corresponding backgrounds. Note sometimes, the defect type can be strongly correlated with the background pattern, which calls for principled method to decouple such interleaved effects.}
        \label{fig:defect_examples}
\end{figure}

Note that defect size generally varies in some large range. In designing the image patch size, as discussed in the preceding subsections, it is necessary to have each patch sufficiently large to contain also a significant background area except a defect, to assure the performance of the self-reference based segmentation. As a result, a defect can be relatively small compared with the patch size. The surrounding background may disturb the classification. In particular we identified an important issue in this stage: the background texture is interleaved with the foreground defect which to some extent affects the learning of classifier. Specifically, for some training samples collected from a certain period and from a certain production line, the defect category distribution may have a strong correlation with the product type during that time. However, this correlation may be due to a short manufacturing system error which may be repaired later. Hence such a correlation becomes a bias in the training data and leads to poor inference performance. In fact, the neural network may be fooled to focus more on the background pattern in the training samples. To address this issue, we propose to add a background channel together with the RGB patch image, in the hope of decoupling the pattern of background and foreground defect. Fig. \ref{fig:defect_examples} shows some typical RGB defect patches. The backgrounds in Fig. \ref{fig:defect_examples} are obtained by the proposed self-reference based method.

More specifically, there are multiple choices for infusing the background information in addition with the defect patch. We consider three cases: i) adding the matched background patch in gray scale  as the fourth channel; ii) adding the binary defect segmentation image as the fourth channel, which implicitly specifies the background in the RGB data; iii) using the three-channel RGB data of the matched background image and combine them into a 6-channel input for DCNN learning. In this paper, we term the above three settings as i) RGB+G, ii) RGB+S, iii) RGB$^2$ respectively, and the vanilla network with three-channel defect patch input is called RGB.

\section{Model Performance Validation}

\textbf{DCNN} We adopted the popular GoogLeNet V3 \cite{inception} for both defect patch identification and defect classification. Although it is a relatively old DCNN in literatures, GoogLeNet has a very good balance in the prediction accuracy level and the computation/resource overhead, thus is very preferred in the industry practice.

\textbf{Experiment Environment} The evaluation is performed on a server with 64G memory, Xeon CPU 2.1GHz and GPU GeForce GTX1080Ti. The software platform includes Ubuntu 16.04, CUDA 8.0, CUDNN 5.1 and Caffe 1.0.

\textbf{Dataset and Protocol} Table \ref{tab:statistics} shows the statistics in terms of the number of training samples and testing samples which are split from the raw data. In fact, we split the raw data into train/test/validate with a ratio 8/1/1. The epoch is set to 100 throughout the experiments. Our evaluation involves two production lines i.e. layer X and layer Y (layer Y is imposed on X), each of which covers three products A, B, C and U, V, W under different textures respectively. Notably, the evaluated real-world data is significantly larger than previous literature in LCD or general panel defect detection, in terms of both defect category number (more than 20) and total sample size (hundreds of thousands of). Examples of defects with different categories are shown in Fig. \ref{fig:defect_examples}. For defect classification, we report the accuracy for each defect category by compared methods which are: i) RGB+G, ii) RGB+S, iii) RGB$^2$ and the vanilla RGB network as described in Sec. \ref{subsec:bdc}.

The inspection images have a uniform size of $1024 \times 768$. We set the size of image patch to $224 \times 224$, which is the same as the GoogLeNet's input image size. For defect patch identification, we adopted a sliding window policy similar to the two-stage R-CNN for candidate generating. The difference is that, we have a strong prior knowledge that each input inspection image mostly contains only one defect area. Thus we only returned the maximum response as the unique defect patch for each image. The sliding step is set to half of the patch size i.e. $112$.

\textbf{Results and Discussion}
In this section, we focus on the performance of the background-aware defect classifier. The evaluation results are reported in Table \ref{tab:results} from which observations can be made.

First, compared with the vanilla RGB network, all the other three models using additional background information obtain improvement. Moreover, the RGB+G model strikes a best tradeoff between efficiency and accuracy. Indeed, the 6-channel RGB$^2$ model is more time costive while the RGB+S which involves a CPU based defect foreground segmentation step is most time consuming (see last row in Table \ref{tab:results} for the average inference time).

Second, we observe the overall classification for layer X is higher than layer Y. This can be explained by the fact that layer X is a more preliminary layer which is imposed by layer Y with more complex textures. Meanwhile, within the same layer, different products have different textures, and the degree of classification difficulty also varies.

Last but not least, there are a few defect types having low accuracy across different layers and products. We note they are very similar to each other. In fact, the defect categorization criteria is inherently based on the manufacturing line root cause other than appearance. This also suggests the limitation of pure appearance based classification system and call for exploration on other information to improve the performance.

\begin{table}[!tb]
\small
        \centering
        \caption{Statistics for different panel layers, products and data splits for training and testing. FAS denotes the particular type for false alarm (no defect) and MISC denotes miscellaneous that do not fall into any other categories. For FAS, it is classified by the locaizer net. Product A, B, C and U, V, W have different textures and X, Y are two panel layers and Y is on the top of X. Note the defect catalogue of layer X and Y are in fact different. We the two layers in fact have different defects category while here they anonymously share the same naming and layer X only has 23 defect categories.}
        \resizebox{0.5\textwidth}{!}{
                \begin{tabular}{l|rr|rr|rr|rr|rr|rr}
                        \toprule
            layer&\multicolumn{6}{c|}{X (anonymous)}&\multicolumn{6}{c}{Y (anonymous)} \\\midrule
            product&\multicolumn{2}{c|}{A (anonymous)}&\multicolumn{2}{c|}{B (anonymous)}&\multicolumn{2}{c|}{C (anonymous)}&\multicolumn{2}{c|}{U (anonymous)}&\multicolumn{2}{c|}{V (anonymous)}&\multicolumn{2}{c}{W (anonymous)}\\\midrule
            split&test&train&test&train&test&train&test&train&test&train&test&train\\\midrule
             type 01&211&1692&423&3384&390&3126&75&600&35&278&141&1128\\
            type 02&465&3720&562&4501&526&4206&182&1460&42&335&122&974\\
            type 03&85&680&455&3642&30&238&592&4741&45&359&594&4751\\
            type 04&232&1854&583&4664&485&3882&65&515&15&120&39&307\\
            type 05&104&830&76&608&46&363&294&2351&73&578&58&468\\
            type 06&198&1586&149&1198&82&656&286&2287&95&766&126&1014\\
            type 07&82&656&140&1120&1&1&589&4717&83&664&595&4755\\
            type 08&73&586&380&3045&285&2279&590&4714&19&156&450&3594\\
            type 09&178&1419&441&3534&308&2466&577&4620&33&266&485&3878\\
            type 10&303&2428&560&4485&429&3431&141&1128&79&630&44&352\\
            type 11&221&1762&84&676&49&394&161&1282&45&365&73&584\\
            type 12&12&90&112&898&75&599&419&3347&3&24&154&1230\\
            type 13&134&1077&205&1635&98&784&79&630&34&271&23&179\\
            type 14&80&639&403&3226&8&58&588&4705&29&234&561&4486\\
            type 15&46&374&376&3007&111&882&106&846&48&378&81&642\\
            type 16&100&794&515&4114&277&2221&131&1042&17&134&93&748\\
            type 17&272&2176&540&4324&439&3514&103&826&47&377&60&478\\
            type 18&51&4034&80&642&28&226&190&1520&53&425&101&807\\
            type 19&575&606&564&4511&724&5792&583&4664&66&523&585&4682\\
            type 20&25&201&255&2046&54&437&301&2408&84&674&100&799\\
            type 21&20&166&103&822&135&1085&260&2075&67&537&205&1634\\
            type 22&---&---&---&---&---&---&144&1146&34&272&80&644\\
            MISC&347&2770&461&3688&332&2652&559&4470&18&148&570&4560\\
            FAS&204&120,000&858&210,000&731&120,000&800&210,000&178&30,000&380&180,000\\
             \bottomrule
                \end{tabular}}\label{tab:statistics}
        \end{table}

\begin{table*}[!tb]
\small
        \centering
        \caption{Accuracy$(\%)$ comparison by different defect types using different input channel settings to the network.}
        \resizebox{1\textwidth}{!}{
                \begin{tabular}{l|rrrr|rrrr|rrrr|rrrr|rrrr|rrrr}
                        \toprule
            layer&\multicolumn{12}{c|}{X (anonymous)}&\multicolumn{12}{c}{Y (anonymous)} \\\midrule
            product&\multicolumn{4}{c|}{A (anonymous)}&\multicolumn{4}{c|}{B (anonymous)}&\multicolumn{4}{c|}{C (anonymous)}&\multicolumn{4}{c|}{U (anonymous)}&\multicolumn{4}{c|}{V (anonymous)}&\multicolumn{4}{c}{W (anonymous)}\\\midrule
            channel&RGB+G&RGB$^2$&RGB&RGB+S&RGB+G&RGB$^2$&RGB&RGB+S&RGB+G&RGB$^2$&RGB&RGB+S&RGB+G&RGB$^2$&RGB&RGB+S&RGB+G&RGB$^2$&RGB&RGB+S&RGB+G&RGB$^2$&RGB&RGB+S\\\midrule
            type 01&\textbf{90.05}&88.63&81.04&85.30&\textbf{86.05}&86.05&83.22&84.63&\textbf{87.18}&86.41&82.31&85.64&\textbf{74.67}&62.67&57.33&56.00&\textbf{94.29}&88.57&88.57&91.43&\textbf{72.34}&71.63&65.96&66.67\\
            type 02&\textbf{86.67}&86.45&82.37&86.02&\textbf{90.21}&89.32&85.23&87.90&\textbf{91.44}&90.68&86.31&88.78&\textbf{78.02}&71.98&72.53&73.63&\textbf{80.95}&71.43&69.05&73.81&\textbf{73.77}&67.21&65.57&68.85\\
            type 03&\textbf{89.41}&87.06&80.00&85.88&\textbf{94.95}&93.63&87.91&90.33&\textbf{80.0}0&73.33&50.00&66.67&\textbf{94.43}&93.41&92.40&93.58&92.24&91.11&86.67&\textbf{93.33}&\textbf{98.15}&96.97&94.78&96.47\\
            type 04&\textbf{94.83}&93.53&86.64&93.53&\textbf{96.57}&96.23&92.80&96.40&\textbf{96.70}&94.64&95.46&95.26&\textbf{83.08}&72.31&63.08&78.46&\textbf{86.67}&73.33&66.67&73.33&\textbf{89.74}&84.62&69.23&82.05\\
            type 05&\textbf{91.35}&89.42&75.00&80.77&\textbf{82.89}&81.58&75.00&78.95&\textbf{93.48}&93.48&80.43&89.13&\textbf{97.96}&97.62&96.60&97.62&\textbf{100}&98.63&97.26&95.89&\textbf{91.38}&91.38&91.38&89.66\\
            type 06&\textbf{96.97}&94.95&90.91&92.42&\textbf{96.64}&94.64&85.91&91.95&\textbf{97.56}&95.12&87.80&95.12&\textbf{95.11}&89.16&89.51&88.81&\textbf{94.74}&93.68&92.63&90.53&87.30&\textbf{89.68}&88.89&87.30\\
            type 07&\textbf{96.34}&92.68&95.12&95.12&\textbf{94.29}&92.86&86.43&92.14&\textbf{100}&100&100&100&\textbf{90.16}&86.25&84.72&86.59&\textbf{83.13}&81.93&77.11&74.70&\textbf{87.56}&87.06&82.18&86.39\\
            type 08&\textbf{78.08}&73.97&68.49&64.38&\textbf{92.31}&90.79&82.63&86.68&\textbf{85.62}&83.15&78.25&76.49&\textbf{77.12}&72.71&73.22&75.59&\textbf{78.42}&67.89&62.63&62.63&\textbf{77.55}&76.88&70.44&74.00\\
            type 09&\textbf{78.65}&78.09&74.16&74.16&\textbf{88.66}&87.30&80.50&79.37&\textbf{80.52}&80.20&78.25&76.62&\textbf{94.28}&91.16&89.25&89.77&84.85&87.88&81.82&\textbf{90.91}&\textbf{91.34}&90.92&86.19&87.22\\
            type 10&88.12&88.12&83.17&\textbf{89.44}&92.32&91.25&87.32&\textbf{92.68}&\textbf{92.07}&91.38&91.38&89.51&\textbf{74.47}&73.76&70.21&68.09&\textbf{88.61}&88.61&83.54&84.81&\textbf{68.18}&65.91&65.91&65.91\\
            type 11&78.73&79.19&\textbf{80.54}&77.38&\textbf{78.57}&69.05&65.48&66.67&\textbf{55.10}&53.06&51.02&46.94&59.63&\textbf{60.87}&58.39&60.25&\textbf{62.75}&60.00&60.00&60.00&\textbf{64.38}&61.64&60.27&61.64\\
            type 12&\textbf{66.67}&50.00&41.67&58.33&\textbf{78.57}&73.32&75.00&77.68&\textbf{94.67}&88.00&77.33&72.00&\textbf{90.69}&89.98&86.16&85.44&\textbf{100}&100&66.67&100&\textbf{85.71}&85.06&83.12&85.06\\
            type 13&\textbf{82.84}&80.60&79.85&76.87&\textbf{78.05}&75.12&75.12&72.68&\textbf{85.71}&84.69&82.65&80.61&68.35&67.09&67.09&\textbf{72.15}&\textbf{91.18}&82.35&88.24&88.24&\textbf{73.91}&69.57&65.22&65.22\\
            type 14&\textbf{88.75}&87.50&81.25&82.50&\textbf{96.53}&96.28&91.81&94.29&\textbf{75.00}&75.00&12.50&37.50&\textbf{95.07}&94.39&86.39&89.97&\textbf{96.55}&93.10&89.66&93.10&\textbf{98.04}&94.12&90.20&90.20\\
            type 15&95.65&94.48&\textbf{97.83}&95.65&96.81&\textbf{97.34}&96.81&96.01&\textbf{93.69}&90.09&89.19&91.89&53.77&54.72&50.00&\textbf{59.43}&\textbf{66.67}&60.42&52.08&56.25&\textbf{55.56}&53.09&49.38&50.62\\
            type 16&63.00&63.00&\textbf{65.00}&62.00&\textbf{78.25}&76.50&76.31&77.09&\textbf{84.12}&83.76&72.92&74.73&87.02&\textbf{87.79}&82.44&87.02&\textbf{89.22}&88.24&70.59&82.35&\textbf{92.47}&91.40&84.95&91.40\\
            type 17&\textbf{88.97}&88.60&80.15&79.04&\textbf{88.15}&85.56&84.44&86.85&\textbf{89.07}&85.87&76.08&74.73&\textbf{35.92}&35.92&28.16&31.07&\textbf{68.08}&61.70&63.83&61.70&30.00&28.33&30.00&\textbf{31.67}\\
            type 18&\textbf{78.43}&74.51&70.59&66.67&\textbf{76.25}&71.25&67.50&65.00&60.00&60.71&46.43&\textbf{79.50}&\textbf{55.79}&54.21&53.16&51.58&\textbf{67.92}&64.15&58.49&58.49&\textbf{53.46}&51.49&51.49&52.47\\
            type 19&\textbf{88.00}&87.13&81.39&82.61&\textbf{77.13}&77.13&74.65&77.13&\textbf{83.98}&82.60&80.52&57.14&\textbf{67.75}&66.21&61.23&59.52&50.00&50.00&50.00&\textbf{51.52}&\textbf{64.45}&60.17&55.56&58.29\\
            type 20&\textbf{72.00}&68.00&52.00&48.00&\textbf{80.78}&80.00&76.08&74.51&\textbf{94.45}&59.26&61.11&81.63&\textbf{75.75}&73.09&69.10&69.44&67.86&64.29&66.67&\textbf{71.43}&\textbf{77.00}&77.00&73.00&73.00\\
            type
            21&\textbf{80.00}&70.00&50.00&50.00&\textbf{94.17}&93.20&78.64&83.50&\textbf{91.85}&91.85&89.63&68.52&\textbf{78.46}&73.08&67.69&71.92&\textbf{85.08}&82.09&83.58&74.63&\textbf{73.66}&73.17&68.29&70.27\\
            type
            22&---&---&---&---&---&---&---&---&---&---&---&---&\textbf{72.92}&63.89&57.64&63.89&\textbf{85.29}&73.53&70.06&79.41&\textbf{58.75}&50.00&51.25&57.50\\
            MISC&\textbf{86.96}&84.15&80.98&85.59&\textbf{92.19}&88.07&79.39&86.73&\textbf{84.59}&83.37&79.15&83.38&\textbf{78.00}&76.57&72.27&77.64&\textbf{77.78}&72.22&55.56&61.11&\textbf{84.56}&78.60&77.72&79.12\\
            FAS&93.13&93.13&93.13&93.13&100&100&100&100&100&100&100&100&99.75&99.75&99.75&99.75&100&100&100&100&99.21&99.21&99.21&99.21\\
            \midrule
            accuracy&\textbf{87.28}&85.99&82.01&83.40&\textbf{89.74}&88.72&85.12&86.93&\textbf{89.79}&87.93&84.40&85.63&\textbf{84.13}&81.86&79.03&80.70&\textbf{83.98}&82.85&79.71&81.24&\textbf{83.50}&81.32&78.00&79.91\\
            time
            (ms)&228.39&235.34&180.09&287.38&247.03&252.02&204.36&293.14&276.22&277.70&234.67&327.86&232.09&233.66&186.51&281.64&237.94&243.74&189.42&290.73&221.72&233.52&179.95&271.93\\
             \bottomrule
                \end{tabular}}\label{tab:results}
        \end{table*}

\section{Visual Inspection Service Architecture}

\begin{figure}[!t]
        \centering
        \includegraphics[width=0.44\textwidth]{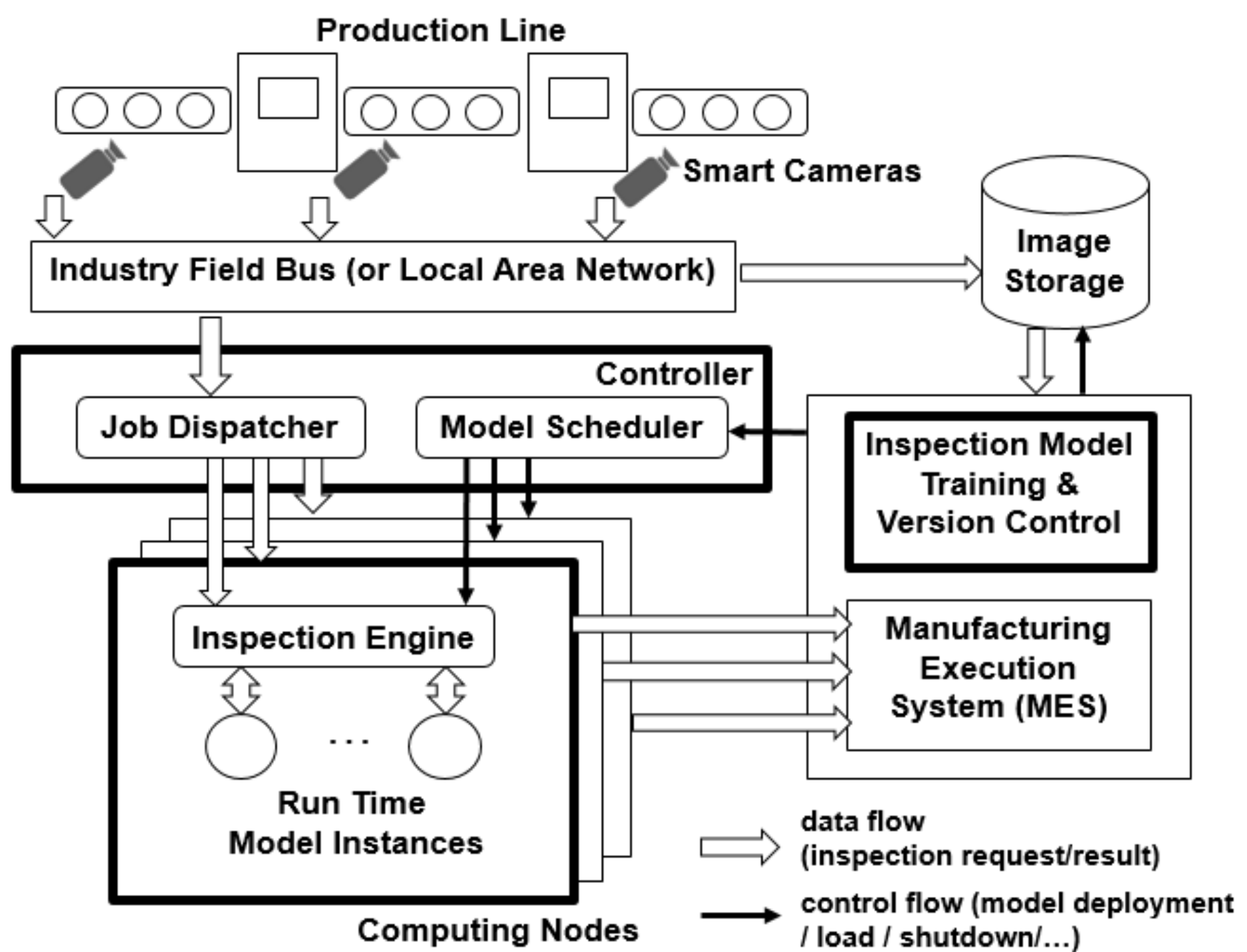}
        \caption{Cognitive visual inspection service architecture}
        \label{fig:archi}
\end{figure}

The architecture of the cognitive visual inspection service is illustrated in Fig.~\ref{fig:archi}. Three main components of visual inspection service are highlighted with bold boxes. In a bottom-up order, the computing nodes are a cluster of GPU-equipped commodity servers that execute the visual inspection tasks. Because the inspection requirements are diverse for different LCD product models and at different stages of production process, different inspection models need to be developed and deployed. In each computing node, an inspection engine takes care of all the local model instances, including model instantiation, model shutdown, inspection task distribution and etc. Inspection requests together with inspection images are routed by a controller (more specifically, the job dispatcher) from front end smart cameras to respective computing nodes. After the request processing, the inspection results are typically sent to the manufacturing execution system (MES). The controller further includes a model scheduler, which sends commands to computing nodes to deploy/load/shutdown models according to dynamic workload, production plans of MES or model update plan. The third component takes charge of model management, that is, the inspection model training and model version control.

The cognitive visual inspection service architecture has been implemented and deployed in a leading LCD manufacturer. The practical operation has witnessed a highly efficient resource utilization (GPU/CPU), easy migration and up-scaling capability as well as sustainable model re-training and extension.

\section{Conclusion}

We propose an automatic TFT-LCD defect detection system in this paper, which is based on a high-performance visual inspection service architecture accommodating complicated AI-based visual inspection models and their speical resource utilization requirements. The visual inspection model is consisted of three key components: defect patch identification, self-reference based defect segmentation and impact evaluation, and background-aware defect classification. In the first component, the weakly supervised learning and periodic patern based approach are jointly utilized to semi-automatically label the defect and defect-free image patches divided from raw training images. On this basis, a binary image patch classifier can be efficiently trained for the defect patch identification. The self-reference based defect segmentation is able to get a good match of defect-free background patch on the same inspection image for each identified defect patch which results in very good defect segmentation effect. The method requires neither pixel-level image annotation nor off-line collected template image. It is very robust to varying image quality that is very common in production lines. Later, in the classification stage, we carefully design an input-channel augmented network to address the challenge of more than 20 classes of defects from real-world data based on the GoogLeNet. In the classification network, the background of the defect area is used as the fourth channels to enrich the input data information and improve the defect classification performance.

We believe our LCD defect detection technology and real-world business experience can also be generalized to other types of panels. There are also directions worth further study:

\textbf{Alleviating Cold-Start by Domain Adaption} As a learning based model, our approach calls for a reasonable size of class-labeled data to learn the classification model, while domain adaption can help re-purpose the learned model from one dataset and category to a new one, given the new domain has only a few data for example when a new product is launched. We note a recent work along this direction \cite{SindagiIJCV17}.

\textbf{Incremental Learning of New Defects} Another open problem is the development of learning new concepts e.g. new defect categories rising from the manufacturing system over time. Moreover, open-set CNN models can also be a relevant solution \cite{BendaleCVPR16}.

\bibliographystyle{IEEEtran}
\bibliography{bare_conf}\ 

\end{document}